\documentclass[manuscript, nonacm]{acmart}

\makeatletter                   
\def\mdseries@tt{m}             
\makeatother                    
\usepackage[plain]{fancyref}
\usepackage[draft=true]{minted} 
\usepackage{color}
\usepackage{hyperref}           
\hypersetup{
    colorlinks=true,
    linkcolor=blue,
    filecolor=red,      
    urlcolor=magenta,
    breaklinks=true,            
}
\usepackage{breakurl}           

\setcopyright{none}

\usepackage{enumitem}
\usepackage{booktabs}
\usepackage{subcaption}

\usepackage{savesym}
\savesymbol{pdfbookmark}
\usepackage{hyperref}
\restoresymbol{HR}{pdfbookmark}

\title{Creating Knowledge Graphs for Geographic Data on the Web}

\author{Elena Demidova}
\affiliation{
  \institution{Data Science and Intelligent Systems Group (DSIS), University of Bonn}
  \city{Bonn}
  \country{Germany}}
\email{demidova@cs.uni-bonn.de}

\author{Alishiba Dsouza}
\affiliation{
  \institution{Data Science and Intelligent Systems Group (DSIS), University of Bonn}
  \city{Bonn}
  \country{Germany}}
\email{dsouza@cs.uni-bonn.de}

\author{Simon Gottschalk}
\affiliation{
  \institution{L3S Research Center, Leibniz University Hannover}
  \city{Hannover}
  \country{Germany}}
\email{gottschalk@L3S.de}

\author{Nicolas Tempelmeier}
\affiliation{
  \institution{L3S Research Center, Leibniz University Hannover}
  \city{Hannover}
  \country{Germany}}
\email{tempelmeier@L3S.de}

\author{Ran Yu}
\affiliation{
  \institution{Data Science and Intelligent Systems Group (DSIS), University of Bonn}
  \city{Bonn}
  \country{Germany}}
\email{ran.yu@cs.uni-bonn.de}

\makeatletter
\let\@authorsaddresses\@empty
\makeatother

\newcommand\blfootnote[1]{%
  \begingroup
  \renewcommand\thefootnote{}\footnote{#1}%
  \addtocounter{footnote}{-1}%
  \endgroup
}

\begin{abstract}
\textbf{Abstract.} Geographic data plays an essential role in various Web, Semantic Web and machine learning applications. OpenStreetMap and knowledge graphs are critical complementary sources of geographic data on the Web. However, data veracity, the lack of integration of geographic and semantic characteristics, and incomplete representations substantially limit the data utility. Verification, enrichment and semantic representation are essential for making geographic data accessible for the Semantic Web and machine learning.
This article describes recent approaches we developed to tackle these challenges. 
\end{abstract}

\begin{document}

\maketitle

\blfootnote{\textcopyright Elena Demidova, Alishiba Dsouza, Simon Gottschalk, Nicolas Tempelmeier, Ran Yu, 2022. 
This is the author's version of the work. It is posted here for your personal use. Not for redistribution. 
The definitive version was published in the ACM SIGWEB Newsletter,  Issue Winter 2022 Article No.: 4 pp 1–8 \url{https://doi.org/10.1145/3522598.3522602}.\\}

\section{Introduction}

Geographic data plays an essential role in a range of real-world applications on the Web, including machine learning models estimating travel time or charging demand for electric vehicles, recommending points of interest and predicting traffic accidents (e.g., \cite{DBLP:conf/itsc/DadwalFD21}, \cite{DBLP:conf/itsc/SaoTD21}). Such applications rely on rich representations of a variety of geographic entities including monuments, roads and charging stations. 

Geographic Web Information Sources and Knowledge Graphs are major complementary Web sources providing information regarding geographic entities, their spatio-temporal context, characteristics, and relationships.

\begin{itemize}[label=---] 
\item \textit{Geographic Web Information Sources} such as OpenStreetMap 
(OSM\footnote{OpenStreetMap, OSM and the OpenStreetMap magnifying glass logo are trademarks of the OpenStreetMap Foundation, and are used with their permission. We are not endorsed by or affiliated with the OpenStreetMap Foundation.}) 
provide characteristics of geographic entities and their relationships. Today, OSM is an essential source of free and open geographic Web information created by voluntary effort, containing over 6.8 billion entities from 188 countries\footnote{\url{https://osmstats.neis-one.org}}.
\item \textit{Knowledge Graphs} such as Wikidata, DBpedia and EventKG \cite{DBLP:journals/semweb/GottschalkD19} contain real-world entities (e.g., persons and places), events and their relationships in a graph-based format. Semantic interpretation of these facts
is facilitated through ontologies. 
\end{itemize}

These data representation paradigms have different focuses: On the one hand, knowledge graphs provide rich semantic information about real-world entities, facilitating querying, exploration, and reasoning. On the other hand, OSM provides rich geographic information, i.e., fine-grained coordinates of real-world locations, but does not possess a clear schema, thus lacking direct semantic interpretation. 

Table~\ref{tab:osm_wikidata_example} illustrates an example geographic entity (``Zugspitze'', a mountain in Germany) and its different representations in OSM and Wikidata\footnote{\texttt{wd} and \texttt{wtd} are the prefixes of \url{http://www.wikidata.org/entity/} and \url{http://www.wikidata.org/prop/direct/}, respectively.}. While OSM provides the information in the form of heterogeneous key-value pairs, so-called ``tags'', e.g., \texttt{natural=peak}, entities in Wikidata are represented by Uniform Resource Identifier (URIs) and are connected to other entities via well-defined properties. 

\begin{table}[ht]
    \caption{Representations of the German mountain Zugspitze in OpenStreetMap and Wikidata. \texttt{wd:Q3375} identifies the Zugspitze in Wikidata.}
    \begin{subtable}{.28\linewidth}
      \centering
        \begin{tabular}{@{}ll@{}}
        \toprule
        \multicolumn{1}{l}{\textbf{Key}} & \multicolumn{1}{c}{\textbf{Value}} \\ \midrule
        name & Zugspitze \\
        natural & peak \\
        summit:cross & yes \\
        ele & 2962 \\ \bottomrule
        \end{tabular}
        \caption{OpenStreetMap representation}
        \label{tab:osm_wikidata_example-osm}
    \end{subtable}%
    \begin{subtable}{.7\linewidth}
      \centering
        \begin{tabular}{@{}lll@{}}
        \toprule
        \multicolumn{1}{c}{\textbf{Subject}} & \multicolumn{1}{c}{\textbf{Predicate}} & \textbf{Object} \\ \midrule
        \texttt{wd:Q3375} & \texttt{rdfs:label} (\textit{label}) & Zugspitze \\
        \texttt{wd:Q3375} & \texttt{wdt:P279} (\textit{instance of}) & mountain \\
        \texttt{wd:Q3375} & \texttt{wdt:P2044} (\textit{elevation}) &  2,962.06 metre \\
        \texttt{wd:Q3375} & \texttt{wdt:P3137} (\textit{parent peak}) & \texttt{wd:Q15127} (\textit{Finsteraarhorn}) \\ \bottomrule
        \end{tabular} 
        \caption{Wikidata representation}
    \end{subtable} 
    \label{tab:osm_wikidata_example}
\end{table}

Knowledge graphs that include data from geographic Web information sources have a significant potential to make geographic data more accessible for the Semantic Web and machine learning applications. 
However, several significant challenges have to be addressed, when it comes to creating such knowledge graphs. 
In particular, this includes:

\begin{enumerate}[label=(C\arabic*)]
    \item Data quality verification: Geographic Web Information Sources, including OSM, often come as volunteered geographic information (VGI), i.e., volunteers contribute to the data collection and modeling. While volunteered contributions increase worldwide coverage of geographic entities, the openness of these sources potentially leads to data quality issues, such as vandalism and misinformation. Therefore, data quality verification is an essential prerequisite for using VGI in knowledge graphs.
    \item Semantic enrichment on schema and data level: Alignments of geographic entities and their classes between geographic Web information sources and knowledge graphs are rare. Consequently, there is a need to establish entity and schema links across sources. The varying coverage and heterogeneous representations of geographic entities, attributes, and relationships make such enrichment particularly challenging.
    \item Creating meaningful representations of geographic entities for semantic and machine learning applications: The combined utilization of Geographic Web Information Sources and knowledge graphs for machine learning requires representations of geographic entities that reflect their semantic and spatial extent.
\end{enumerate}

\begin{figure}[ht]
  \centering
  \includegraphics[width=\textwidth]{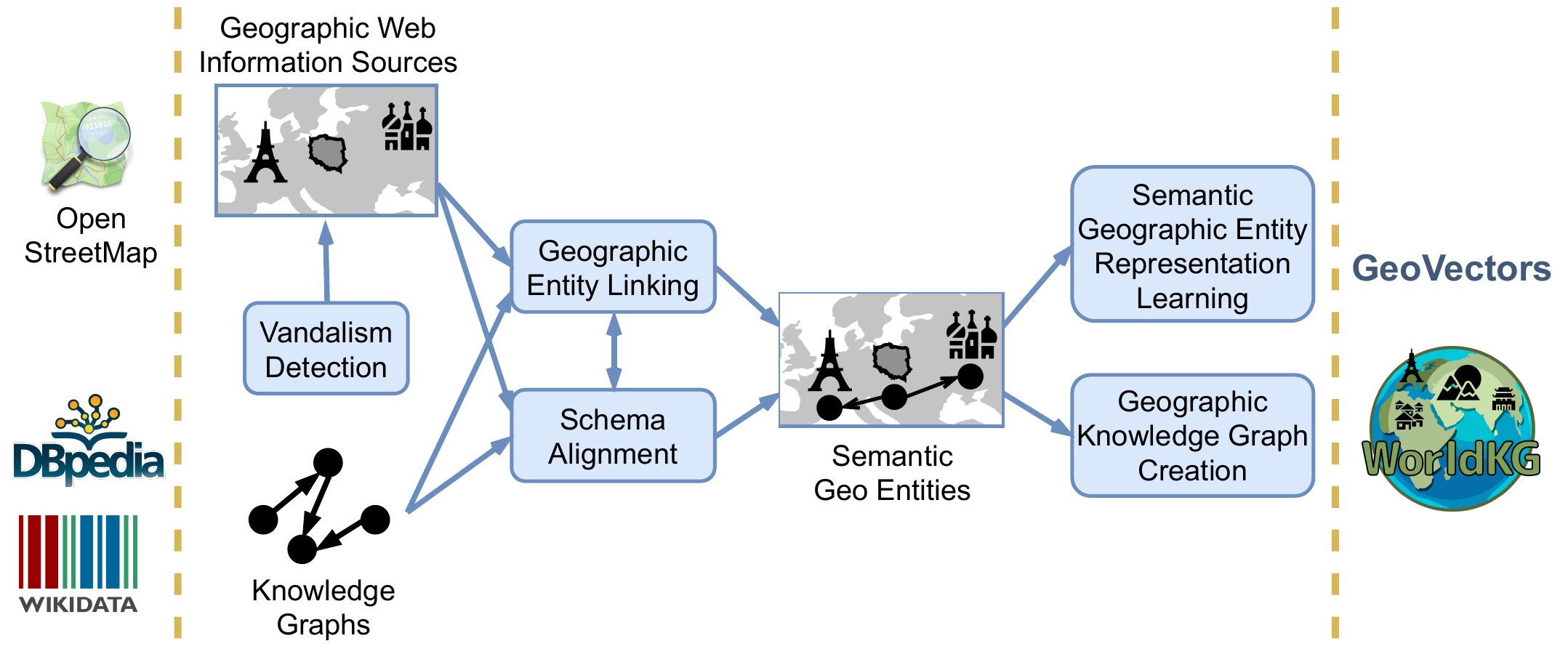}
  \caption{Overview of the resources and methods for creating knowledge graphs for geographic data on the Web discussed in this article.}
  \label{fig:overview}
\end{figure}

Recently, we developed several methods to tackle these challenges, making geographic entities available for downstream applications.
An overview of our methods is shown in Fig.~\ref{fig:overview}. We consider existing knowledge graphs and OSM as input data sources. To enable data quality verification (C1), we propose a supervised method for \emph{vandalism detection} in OSM that can cope with crowdsourced geographic Web data. To tackle (C2), we establish links between geographic entities in OSM and a knowledge graph through \emph{geographic entity linking}. Based on interlinked geographic entities, \emph{schema alignment} aims at aligning their classes (e.g., ``mountain'' in Wikidata and ``peak'' in OSM). Finally, we make the resulting semantic geographic entities available in two forms: (i) as a WorldKG knowledge graph through \emph{geographic knowledge graph creation}, and (ii) through \emph{semantic geographic entity representation learning}, to represent geographic entities for the usage in machine learning applications (C3).

In the remainder of this article, we provide more details of these methods and the datasets built upon them. First, we present our methodology: Section~\ref{sec:ovid} discusses the \emph{OVID} approach for OSM vandalism detection, Section~\ref{sec:osm2kg} presents \emph{OSM2KG} for linking Wikidata and OSM entities, and Section~\ref{sec:nca} introduces the \emph{NCA} approach for schema alignment between Wikidata, DBpedia and OSM. Then, we describe the creation of two resources: \emph{WorldKG} presented in Section~\ref{sec:worldkg} is our geographic knowledge graph, and \emph{GeoVectors} presented in Section~\ref{sec:geovectors} is our resource of geographic entity embeddings covering their spatial and semantic aspects. Finally, in Section~\ref{sec:future_work}, we provide a conclusion and discuss future work.


\section{Vandalism Detection -- OVID}
\label{sec:ovid}

Ensuring a high quality of openly available geographic information is vital for adopting this information in real-world applications. 
In particular, protecting crowdsourced geographic data from vandalism is a challenging task.
Whereas this task has been addressed in knowledge bases such as Wikipedia and Wikidata, existing vandalism detection approaches typically rely on textual features and do not consider the spatial dimension of the data. 

In \cite{tempelmeier2022}, we developed novel methods to automatically detect vandalism on OpenStreetMap.
We proposed the OVID (\underline{O}penStreetMap \underline{V}andal\underline{i}sm \underline{D}etection) model.
OVID combines user, edit, and changeset features with an attention-based neural architecture in a supervised classification model.
Furthermore, we systematically extracted vandalism incidents from the OSM edit history from 2014 to 2019 and provided a dataset containing over 9,000 real-world vandalism examples.
We made the dataset available under an open license\footnote{\url{https://github.com/NicolasTe/Ovid}}.
Our evaluation results confirmed that the OVID model outperformed the existing models for vandalism detection.


\section{Geographic Entity Linking -- OSM2KG}
\label{sec:osm2kg}

According to the five-star open data principle proposed by Tim Berners-Lee, the interlinking across open datasets is an essential data quality indicator.
However, such entity links are rarely available in geographic Web information sources.
For instance, as of January 2021, only 0.05\% of all OSM entities provided a link to the Wikidata knowledge graph.

To systematically increase the number of links between OSM and knowledge graph entities, we developed the \emph{OSM2KG} approach \cite{TEMPELMEIER2021349}.
Given an OSM entity, OSM2KG aims to determine a geographic entity in a knowledge graph representing the same real-world entity.
In contrast to a knowledge graph, OSM does not enforce a strict schema and represents its entities through key-value pairs (see Table \ref{tab:osm_wikidata_example-osm}). This representation leads to sparse attributes and imposes significant challenges to feature extraction for traditional link discovery. 
OSM2KG is a supervised machine learning approach for linking OSM and knowledge graphs at the entity level. 
The core of OSM2KG is a novel key-value embedding for OSM entities. 
OSM2KG applies an unsupervised representation learning algorithm that captures OSM entity semantics. The resulting embedding represents OSM entities in a classification model for link discovery.
Our evaluation conducted on OpenStreetMap and the Wikidata and DBpedia knowledge graphs demonstrates that OSM2KG reliably outperforms existing link discovery approaches.
We make our code available under an open license\footnote{\url{https://github.com/NicolasTe/osm2kg}}.

\section{Schema Alignment -- NCA}
\label{sec:nca}

Although OpenstreetMap contains numerous geographic entities, these entities are not directly accessible to semantic applications due to their key-value-based structure, as discussed in Section~\ref{sec:osm2kg}. Aligning schema elements such as classes and properties of knowledge graphs to OSM keys and key-value pairs (tags) can benefit semantic applications, providing direct access to geographic entities on a world scale. 
Existing approaches for schema alignment are typically based on structural similarity, or string similarity \cite{NgoBT13,madhavan2001generic}.
Due to its flat schema and heterogeneous keys, such methods cannot be applied to OSM. 
To tackle these challenges, we introduced a two-step approach called NCA (\underline{N}eural \underline{C}lass \underline{A}lignment) for the tag-to-class alignment between OSM and knowledge graphs~\cite{iswcNCA}.

In the first step, NCA utilizes existing links between OSM and knowledge graphs as a supervision signal to classify OSM entities into the semantic classes of knowledge graphs. As a result of the classification, a shared latent space capturing the semantics of tag-to-class alignment is created. NCA systematically probes the classification model with specific tags in the second step to obtain the corresponding semantic classes. We made our code available under an open license\footnote{\url{https://github.com/alishiba14/NCA-OSM-to-KGs}}.

We evaluated the NCA approach on six country-specific datasets of different countries using manually annotated tag-to-class pairs. 
We observed that NCA outperforms state-of-the-art approaches. As the result of this alignment, we also observed an over 400\% increase in the semantic class annotations for OSM data. %
NCA mapping of the schema elements is an essential step towards the semantic representation of OSM entities. 
This schema mapping enables us to enrich OSM entities with semantic class information using knowledge graphs. We utilize the tag-to-class matches obtained by NCA 
for the construction of the WorldKG knowledge graph, a new geographic knowledge graph built starting from OSM, presented in Section ~\ref{sec:worldkg}.

\section{Geographic Knowledge Graph Creation -- WorldKG}
\label{sec:worldkg}

Knowledge graphs are vital sources of semantic information regarding real-world entities and their relations. 
However, the coverage of geographic entities in popular knowledge graphs is relatively poor. 
As discussed in Section~\ref{sec:osm2kg}, on the one hand, popular general-purpose knowledge graphs such as DBpedia and Wikidata only provide a limited number of geographic entities. 
On the other hand, specialized geographic knowledge graphs such as LinkedGeoData \cite{auer2009linkedgeodata} and YAGO2geo \cite{karalis2019extending} contain only a tiny subset of geographic classes. 
To bridge this gap, we develop a new geographic knowledge graph \emph{WorldKG} \cite{dsouza2021worldkg}, aiming to provide better coverage of semantic representations of geographic entities through a fusion of knowledge graphs and OpenStreetMap.

The WorldKG knowledge graph\footnote{\url{https://www.worldkg.org/}} currently contains over 100 million geographic entities from 188 countries. 
This knowledge graph is based on a novel WorldKG ontology having over 1000 geographic classes created using the OSM schema. 
We convert the flat OSM schema into a hierarchical ontology structure using the schema alignment NCA approach presented in Section \ref{sec:nca}. 
The results of this alignment can determine the correct classes of WorldKG entities in Wikidata and DBpedia ontology with over 99\% accuracy, on average. 

To enable direct semantic access to WorldKG, we provide a SPARQL endpoint and downloadable data files in standard RDF turtle format. We believe that the scale and accuracy of WorldKG can help various semantic data-driven applications. 
Examples include event-centric and geospatial question answering and geographic information retrieval.

\section{Semantic Geographic Entity Representation Learning -- GeoVectors}
\label{sec:geovectors}

Usable representations of geographic entities are of utmost importance for various applications, including travel time estimation, location recommendation, and geographic information retrieval. Dependent on the specific application, such representations need to capture the spatial and semantic entity dimensions.
The spatial dimension reflects the geographic extent and proximity to neighboring entities. The semantic dimension can include an entity type and further type-dependent attributes.
In OSM, where geographic entities are represented using key-value pairs reflecting various entity types and their heterogeneous attributes, discrete vector-based entity representations are extremely sparse and high-dimensional. 
The computation of geographic entity representations that machine learning algorithms can effectively use is not trivial.

In our \emph{GeoVectors} corpus \cite{10.1145/3459637.3482004}, we provide a ready to use corpus containing embeddings of over 980 million geographic entities extracted from OpenStreetMap.
GeoVectors corpus consists of two parts. 
The \emph{GeoVectors-location} embeddings capture the spatial dimension of the geographic entities.
To this extent, we adopted an established geographic representation learning algorithm and developed a framework that enables the algorithm training on a world scale.
The \emph{GeoVectors-tags} embeddings capture the semantic dimension of geographic entities.
To obtain these embeddings, we adopted a pre-trained word embedding model to map OSM key-value pairs into a latent space and obtain the semantic representation of geographic entities.
Our experiments conducted on real-world data demonstrate that the GeoVectors embeddings effectively capture the spatial and semantic dimensions of geographic entities.

We offer direct access to the GeoVectors corpus on our website\footnote{\url{https://geovectors.l3s.uni-hannover.de}}.
On this website, we provide a detailed corpus description.
Moreover, we provide a knowledge graph that integrates GeoVectors with popular knowledge graphs such as Wikidata and DBpedia, making our embeddings of geographic entities in these sources directly accessible.
We make our knowledge graph accessible via a SPARQL endpoint.
Finally, we provide a search function that enables name-based entity access to the GeoVectors knowledge graph.

\section{Conclusion and Future Work}
\label{sec:future_work}

In this article, we described three significant challenges towards the creation of knowledge graphs for geographic data on the Web, including data quality verification, semantic enrichment, and the creation of effective representations. We tackled these challenges with several new approaches, including OVID for vandalism detection, OSM2KG for geographic entity linking, and NCA for schema alignment. These approaches build the basis for creating the WorldKG knowledge graph for geographic data and the GeoVectors corpus containing distributional representations of geographic entities.

In future work, we aim to provide an increasingly complete and semantically rich representation of geographic data on the Web.
We will further work on bringing these methods closer together to enable cross-fertilization. Furthermore, WorldKG and GeoVectors can benefit from interlinking and enrichment with additional datasets. We make the datasets openly available to enable their broad reuse.

\section*{ACKNOWLEDGMENTS}
This work was partially funded by DFG, German Research Foundation (``WorldKG'', 424985896), the Federal Ministry for Economic Affairs and Climate Action (BMWK), Germany (``d-E-mand'', 01ME19009B and ``CampaNeo'', 01MD19007B), the Federal Ministry of Education and Research (BMBF), Germany (``Simple-ML'', 01IS18054), and the European Commission (EU H2020, ``smashHit'', grant-ID 871477).

\bibliographystyle{ACM-Reference-Format}

\bibliography{references}







   
\end{document}